\pdfoutput=1
\documentclass[11pt]{article}

\usepackage{acl}

\usepackage{times}
\usepackage{latexsym}
\usepackage{comment}
\usepackage{color}
\usepackage{makecell}
\usepackage{colortbl}
\usepackage[T1]{fontenc}
\usepackage[utf8]{inputenc}

\usepackage{microtype}
\usepackage[most]{tcolorbox}
\usepackage{inconsolata}

\usepackage{mathtools}
\usepackage{makecell}

\usepackage[most]{tcolorbox}
\tcbset{
  colback=white,
  colframe=black!50,
  boxrule=0.5pt,
  arc=2pt,
  left=6pt,
  right=6pt,
  top=6pt,
  bottom=6pt
}
\newtcbtheorem[]{mydef}{Definition}%
{colback=white!98!black,colframe=black!15!black,fonttitle=\bfseries}{def}
\usepackage{array,multirow,graphicx}
\usepackage{amsmath}
\usepackage{amssymb}
\usepackage{booktabs}
\usepackage{algpseudocode}

\usepackage[export]{adjustbox}
\usepackage{float}

\usepackage[capitalize]{cleveref}
\crefname{section}{Sec.}{Secs.}
\Crefname{table}{Table}{Tables}
\usepackage{amsmath,amsfonts,bm}

\def\eqref#1{equation~\ref{#1}}
\def\1{\bm{1}}

\DeclareMathAlphabet{\mathsfit}{\encodingdefault}{\sfdefault}{m}{sl}
\SetMathAlphabet{\mathsfit}{bold}{\encodingdefault}{\sfdefault}{bx}{n}

\usepackage{geometry}
\usepackage{booktabs} 
\usepackage{bbm}
\usepackage{mathtools}
\usepackage{nccmath}
\usepackage{setspace}

\usepackage{caption}
\usepackage{subcaption}

\usepackage[linesnumbered,ruled,vlined]{algorithm2e}

\SetKwInput{KwInput}{Input}                % Set the Input
\SetKwInput{KwOutput}{Output}              % set the Output

\SetCommentSty{mycommfont}

\SetAlCapSty{algcapsty}

\usepackage[T1]{fontenc}
\usepackage{wrapfig,lipsum,booktabs}

\usepackage{soul}
\usepackage{dsfont}
\usepackage{enumerate}
\usepackage{enumitem}

\usepackage{amsmath}
\usepackage{amsfonts}
\usepackage{bbm}
\usepackage{dsfont}
\usepackage[Symbol]{upgreek}
\usepackage{lscape}
\usepackage{caption}
\usepackage{balance}
\usepackage{xspace}
\usepackage{float}
\usepackage{kotex}

\usepackage{wasysym}
\usepackage{xcolor}
\usepackage{multirow}
\usepackage{array, boldline, rotating}

\usepackage{amssymb}% http://ctan.org/pkg/amssymb
\usepackage{pifont}% http://ctan.org/pkg/pifont
\renewcommand*\eqref[1]{(\ref{#1})}

\newcommand{\yes}[1]{\textcolor{blue}{[YES]}}
\newcommand{\no}[1]{\textcolor{orange}{[NO]}}
\newcommand{\na}[1]{\textcolor{gray}{[N/A]}}
\newcommand{\eg}{\emph{e.g.,~}}
\newcommand{\ie}{\emph{i.e.,~}}

\newcommand{\myparagraph}[1]{\vspace{0.07cm}\noindent\textbf{#1}~}

\definecolor{LightCyan}{rgb}{0.88,1,1}
\definecolor{Blue}{rgb}{0, 0.5, 1}
\definecolor{Green}{rgb}{0.0, 0.8, 0.0 }
\definecolor{Red}{rgb}{0.95, 0.55, 0.6}
\definecolor{Skyblue}{rgb}{0.6, 0.6, 0.95 }

\NewDocumentEnvironment{suptitle}{ +b }{
    \twocolumn[{#1}]%
}{}

\NewDocumentCommand{\supptitle}{s}{
\begin{suptitle}
        \centering
        \rule{\textwidth}{0.03cm}\\[0.1cm]
        -Supplementary Material-\\[0.2cm]
        {\Large 
            \textbf{\mytitle }
        }\\%[0.40cm]
        \rule{\textwidth}{0.03cm}\\[0.2cm]
\end{suptitle}}

\NewDocumentCommand{\summarytitle}{s}{
\begin{summarytitle}
        \centering
        \rule{\textwidth}{0.03cm}\\[0.1cm]
        -Summary of Changes-\\[0.2cm]
        {\Large 
            \textbf{\mytitle }
        }\\%[0.40cm]
        \rule{\textwidth}{0.03cm}\\[0.2cm]
\end{summarytitle}}

\makeatletter
\long\def\@makefntext#1{%
  \noindent\@makefnmark\ #1
}
\makeatother

\newcommand{\llama}{Llama}
\newcommand{\myalg}{\textbf{\texttt{PatchRAG}}\xspace}

\newcommand{\mytitle}{Feedback Adaptation for Retrieval-Augmented Generation}
\title{\mytitle}
\author{Jihwan Bang$^1$\hspace{1em}Seunghan Yang$^1$\hspace{1em}Kyuhong Shim$^{2}\footnotemark[1]$\hspace{1em}Simyung Chang$^{3}\footnotemark[1]$\\\textbf{Juntae Lee$^1$}\hspace{1em}\textbf{Sungha Choi$^{4}$\footnotemark[1]\footnotemark[2]}\\
{$^1$Qualcomm AI Research\footnotemark[3], Qualcomm Korea YH, Seoul, Republic of Korea} \\ 
{$^2$Sungkyunkwan University}\hspace{2em}{$^3$H1R.ai}\hspace{2em}{$^4$Kyung Hee University} \\
{\texttt {\small\{jihwbang, seunghan, juntlee\}@qti.qualcomm.com}}\hspace{2em}{\texttt {\small khshim@skku.edu}} \\
{\texttt {\small simyung@h1r.ai}}\hspace{2em}{\texttt {\small sunghac@khu.ac.kr}}
}
\begin{document}
\maketitle
\renewcommand{\thefootnote}{\fnsymbol{footnote}} % 기호 footnote로 변경
\footnotetext[1]{Work completed while employed at Qualcomm AI Research.}
\footnotetext[2]{Corresponding author.}
\footnotetext[3]{Qualcomm AI Research is an initiative of Qualcomm Technologies, Inc.}

\begin{abstract}
Retrieval-Augmented Generation (RAG) systems are typically evaluated under static assumptions, despite being frequently corrected through user or expert feedback in deployment. Existing evaluation protocols focus on overall accuracy and fail to capture how systems adapt after feedback is introduced.
We introduce \textit{feedback adaptation} as a problem setting for RAG systems, which asks how effectively and how quickly corrective feedback propagates to future queries. To make this behavior measurable, we propose two evaluation axes: \textit{correction lag}, which captures the delay between feedback provision and behavioral change, and \textit{post-feedback performance}, which measures reliability on semantically related queries after feedback.
Using these metrics, we show that training-based approaches exhibit a trade-off between delayed correction and reliable adaptation. We further propose \myalg, a minimal inference-time instantiation that incorporates feedback without retraining, demonstrating immediate correction and strong post-feedback generalization under the proposed evaluation. Our results highlight feedback adaptation as a previously overlooked dimension of RAG system behavior in interactive settings.
    
\end{abstract}

%\jt{}

\section{Introduction}
\label{sec:intro}

Retrieval-Augmented Generation (RAG) has emerged as a dominant paradigm for grounding large language models~\cite{dubey2024llama, achiam2023gpt, team2024gemma} in external knowledge. By coupling a retriever with a generator, RAG systems~\cite{karpukhin2020dense, yang2025learning, wanginstructretro, linra, sun2023chatgpt} have demonstrated strong performance across question answering and knowledge-intensive tasks. However, existing RAG research largely assumes that the underlying knowledge or system behavior remains static after deployment. In practice, deployed RAG systems are frequently corrected through user or expert feedback when they produce outdated, incorrect, or undesirable outputs. Despite its practical importance, how a RAG system should adapt after receiving feedback remains underexplored and poorly formalized.

Most prior approaches~\cite{siriwardhana2023improving, zhang2024raft, mao2024rag} address feedback by retraining or fine-tuning components of the system, such as the retriever or the generator. While effective in improving accuracy in aggregate, these training-based updates introduce an inherent delay between the moment feedback is received and the moment the system’s behavior actually changes. Moreover, existing evaluation protocols~\cite{petroni2021kilt, trivedi2022musique, thakur2021beir} focus almost exclusively on overall task accuracy, making it difficult to reason about how quickly and reliably feedback influences future predictions. As a result, current RAG benchmarks and methods conflate correctness with adaptability, obscuring a critical dimension of system behavior in real-world deployments.

In this work, we argue that adapting to feedback is a distinct research problem that cannot be adequately studied through standard RAG evaluation. We introduce \textit{feedback adaptation} as a new problem setting for RAG systems, and here is the research question: \textit{once a system is corrected, how effectively and how quickly does this correction propagate to future queries?} Crucially, feedback adaptation is not about improving average accuracy, but about characterizing the dynamics of knowledge updates under interaction.

\begin{figure*}[t]
    \centering
    \includegraphics[width=\linewidth]{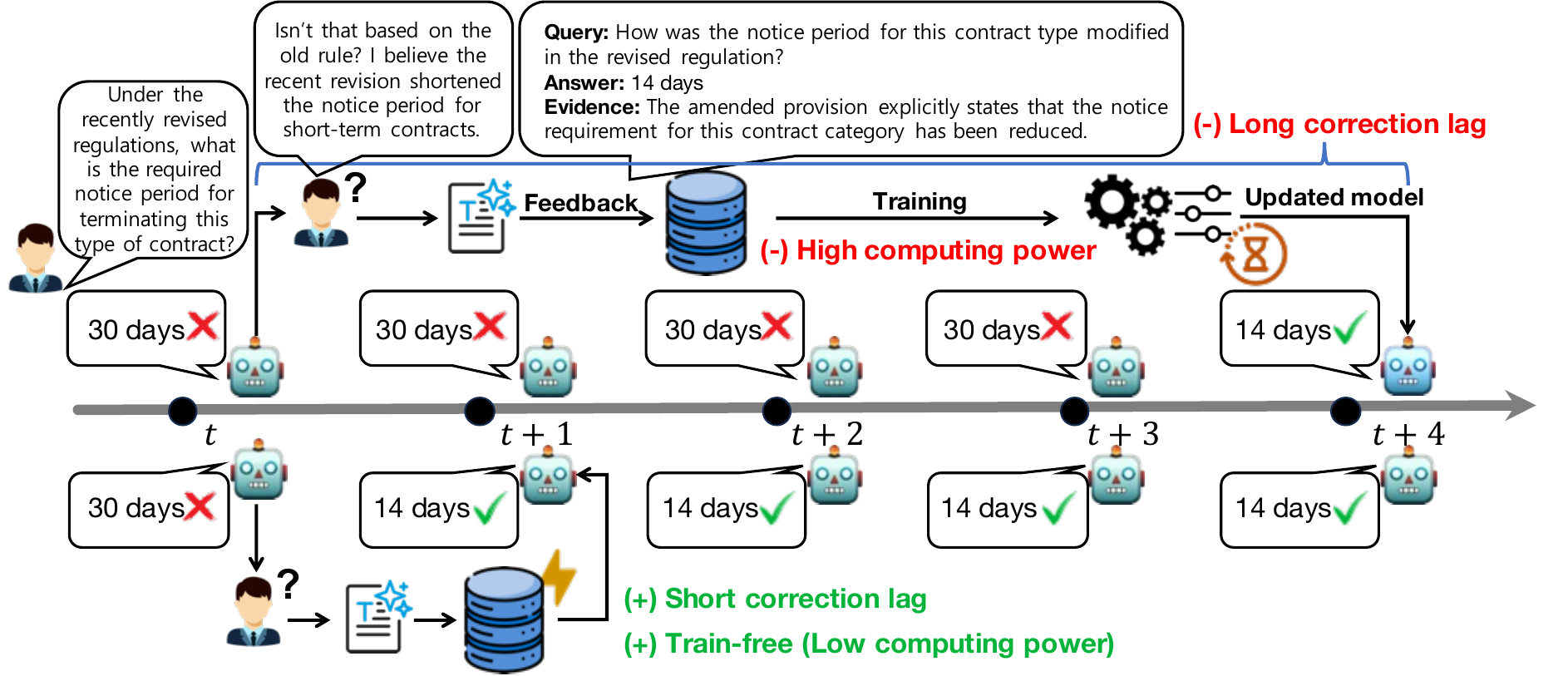}
    \caption{
        \textbf{Illustration of feedback adaptation in deployed RAG systems.}
\textbf{(Upper)} Under training-based update paradigms, feedback is incorporated only after offline or periodic retraining, resulting in substantial correction lag. Even after expert feedback identifies an incorrect answer, the system may continue to reproduce the same error until the update is completed.
\textbf{(Lower)} Inference-time feedback incorporation enables immediate behavioral correction by conditioning generation on stored feedback, achieving low correction lag and reliable generalization to semantically related queries.
}
    \label{fig:concept_comparison}
\end{figure*}

To make this problem explicit and measurable, we formalize two orthogonal evaluation axes. First, we define \textit{correction lag}, which captures the delay between the time feedback is provided and the time at which the system’s outputs consistently reflect that feedback. As illustrated in \autoref{fig:concept_comparison}, different update paradigms induce fundamentally different temporal behaviors after feedback is provided. Any approach that relies on retraining or redeployment necessarily incurs a significant correction lag, regardless of final accuracy. Second, we define \textit{post-feedback performance}, which measures a system’s reliability on queries that are semantically consistent with the provided feedback. Unlike standard test accuracy, this metric isolates the effect of feedback and evaluates whether the system has meaningfully internalized the correction rather than merely memorizing labels.

Together, these two axes expose a trade-off that is largely invisible under existing evaluation protocols. Training-based methods can achieve strong performance, but only after incurring substantial correction lag. Conversely, systems that react immediately to feedback risk shallow or brittle generalization. We show empirically that this trade-off is not an artifact of specific models or datasets, but a structural property of how feedback is incorporated in current RAG paradigms.

To ground the feedback adaptation problem, we present \myalg, a minimal reference instantiation that incorporates feedback at inference time by storing and retrieving feedback patches. \myalg is intentionally simple, serving as a proof-of-concept that immediate adaptation with strong post-feedback performance is possible without retraining. By evaluating \myalg alongside training-based baselines under the proposed metrics, we illustrate behaviors that are obscured by standard accuracy-driven evaluation.

Our contributions are threefold:
(1) we introduce \textit{feedback adaptation} as a new problem setting for RAG systems.
(2) we formalize two evaluation metrics---\textit{correction lag} and \textit{post-feedback performance}---that capture complementary aspects of adaptation; and
(3) we provide a minimal instantiation and empirical study, including stress tests under practically relevant imperfect-feedback conditions, that demonstrates the practical and conceptual implications of this problem formulation.
\section{Feedback Adaptation}
\label{sec:problem}

\subsection{Motivation: Why Existing RAG Evaluation Is Insufficient}

Retrieval-Augmented Generation (RAG) systems are typically evaluated under the assumption that model parameters and retrieval corpora remain fixed during evaluation. Under this setting, performance is measured by aggregate accuracy metrics such as EM or F1 over a held-out test set. However, this evaluation paradigm fails to capture a critical aspect of deployed RAG systems: they are routinely corrected after making mistakes.

When a system produces an incorrect or undesirable output, users or domain experts often provide feedback in the form of corrected answers, supporting evidence, or clarifications. Once such feedback is given, the relevant question is no longer whether the system performs well on average, but whether it adapts its future behavior in response to that correction. Existing RAG benchmarks and metrics are not designed to answer this question, as they do not condition evaluation on the occurrence of feedback.

Crucially, this limitation is not merely an evaluation artifact. It reflects a deeper gap in how RAG systems are conceptualized: adaptation after feedback is implicitly treated as a training or maintenance concern, rather than as a first-class research problem. As a result, different system designs that behave identically under static evaluation can exhibit dramatically different behaviors once feedback is introduced.

\subsection{Feedback Adaptation: Problem Definition}

We formalize \textit{feedback adaptation} as a problem setting for RAG systems that explicitly models how system behavior evolves after feedback. Consider a RAG system that receives a sequence of user queries, $q_u = \{q_1, q_2, \cdots, q_t\}$. 
At time step $t$, the system produces an output $a_t$ for query $q_t$. If the output is incorrect or unsatisfactory, an external agent provides feedback $f_t$, which may include a corrected answer or additional context. A system exhibits feedback adaptation if, after receiving feedback $f_t$, its subsequent outputs reflect that feedback when answering future queries that are semantically consistent with $q_t$. 

Importantly, feedback adaptation does not require memorizing the exact feedback instance. Rather, it requires generalizing the correction to a set of related queries, potentially expressed with different surface forms or contexts. This distinguishes feedback adaptation from supervised learning, continual learning, and model editing, which focus on aggregate performance, retention of past knowledge , or targeted parameter updates, respectively. In contrast, feedback adaptation concerns the \textbf{temporal dynamics of system behavior under interaction}, conditioned on when and how feedback is received.

\subsection{Evaluation Axis I: Correction Lag}
\label{subsec:correction_lag}
The first axis of feedback adaptation is \textit{correction lag}, which measures how quickly feedback influences system behavior. Given feedback $f_t$  at time $t$, correction lag is defined as the elapsed time until the system consistently produces corrected outputs that reflect the feedback for semantically consistent queries. 

This notion captures a fundamental limitation of training-based approaches. Any method that relies on retraining, fine-tuning, or redeployment after feedback necessarily incurs a significant correction lag, regardless of the final accuracy achieved after the update. Correction lag is therefore orthogonal to accuracy: two systems may achieve identical post-update performance, yet differ substantially in how long they continue to produce incorrect outputs after feedback is given. 

\subsection{Evaluation Axis II: Post-Feedback Performance}
\label{subsec:post_feedback}
While correction lag captures when a system adapts, it does not capture how well the system adapts once feedback has taken effect. We therefore introduce \textit{post-feedback performance}.
Post-feedback performance measures a system’s accuracy on queries that are semantically consistent with previously provided feedback, after the system has incorporated that feedback.
Unlike standard test accuracy, this metric explicitly conditions evaluation on the presence of feedback and focuses on intent-consistent generalization. A system that simply memorizes the feedback instance without generalizing to related queries may exhibit low post-feedback performance.

Together with correction lag, post-feedback performance captures whether a system has meaningfully internalized feedback rather than treating it as an isolated exception.

% 2.5 Implications and Design Space

% The two evaluation axes defined above reveal a fundamental trade-off in feedback adaptation. Systems that rely on retraining can eventually achieve strong post-feedback performance, but only after incurring substantial correction lag. Conversely, systems that update immediately may suffer from shallow or brittle generalization if feedback is incorporated naively.

% This trade-off is largely invisible under conventional RAG evaluation, which collapses temporal behavior into a single aggregate score. By making feedback adaptation explicit, we expose a broader design space of RAG systems that differ not only in accuracy, but in how they evolve under interaction.

% In the remainder of this paper, we explore this design space empirically. Rather than proposing a complex new architecture, we introduce a minimal reference instantiation that demonstrates the feasibility and implications of feedback adaptation under the proposed metrics.
\section{\myalg}
\label{sec:patchrag}

This section describes \myalg, a minimal instantiation of feedback adaptation introduced in~\autoref{sec:problem}. \myalg incorporates feedback at inference time by storing and retrieving feedback patches, enabling immediate behavioral updates without retraining. The design avoids architectural modifications to isolate the effect of feedback incorporation under the proposed evaluation axes. As such, \myalg satisfies two requirements implied by feedback adaptation: (\emph{i}) immediate incorporation of feedback, resulting in minimal correction lag; and (\emph{ii}) generalization beyond individual feedback instances, enabling strong post-feedback performance on semantically related queries.

% \subsection{Why Retrieval Alone Is Not Enough}
% \label{subsec:motivation}
% A natural strategy for incorporating expert feedback is to append the corrected evidence $c^*$ to the document corpus and rely on the retriever to surface it for future queries.
% However, this strategy often proves insufficient in practice.
% User queries primarily express \emph{intent}, whereas document chunks, including corrected evidence, encode \emph{declarative content}.
% As a result, dense retrievers frequently prioritize semantically adjacent facts rather than the specific feedback required to correct prior errors.

% This misalignment prevents corrected evidence from reliably influencing model behavior, even after it has been added to the corpus.
% Consequently, the system may continue to produce the same incorrect answers for related queries, leading to \textit{poor post-feedback performance}.
% In the feedback adaptation setting, this failure directly undermines effective adaptation: the corrected information is available, but the retrieval fails to expose it.

\subsection{Intent-Context Retrieval for Feedback Adaptation}
\label{subsec:retrieval}

% To address this limitation from~\autoref{subsec:motivation}, 
\myalg augments standard content-based retrieval with \emph{intent-aware feedback retrieval}.
Each stored feedback item is represented as a tuple
$f_i = (q_i, a_i, c_i)$,
consisting of the original query, the corrected answer, and supporting evidence.

Given a new query $q$, \myalg assigns each feedback item a relevance score
\begin{equation}
\label{eq:patchrag_sim}
    S_i(q)
    = \lambda \cdot \mathrm{sim}(q, q_i) + (1-\lambda) \cdot \mathrm{sim}(q, c_i),
\end{equation}
where $\mathrm{sim}(\cdot,\cdot)$ denotes cosine similarity between embeddings, and $\lambda \in [0,1]$ controls the balance between intent matching and context grounding.

\myparagraph{Intent matching.}
The term $\mathrm{sim}(q, q_i)$ captures similarity at the level of user intent, enabling retrieval of relevant feedback even when the query and corrected evidence share little surface or semantic overlap.

\myparagraph{Context grounding.}
The term $\mathrm{sim}(q, c_i)$ anchors retrieval to content-level relevance, preserving stability for queries whose intent has not been previously observed and preventing spurious matches driven solely by intent similarity.

The top-$k$ feedback items selected by $S_i(q)$ are used as corrective exemplars for generation.

\subsection{Inference with In-Context Learning}
\label{subsec:generation}

\myalg incorporates retrieved feedback directly into generation through in-context learning.
Given the selected feedback items $\{p_{i_1}, \dots, p_{i_k}\}$, we construct an augmented prompt (see~\autoref{sec:prompt_rag} for prompt template) that presents the feedback, followed by the current user query $q$.

This approach allows the generator to immediately condition its predictions on the newly provided feedback without any gradient-based updates or parameter modification.
As a result, feedback can influence subsequent predictions as soon as it is added to memory, yielding low correction lag while maintaining strong post-feedback performance.

\myparagraph{Scope.} \myalg is intentionally minimal. Its purpose is not to serve as a final solution to feedback adaptation, but to provide a concrete reference point within the design space exposed by the proposed evaluation axes. By avoiding retraining, architectural changes, or specialized optimization objectives, \myalg isolates the effect of feedback incorporation itself. 
\section{Experiments}
\label{sec:experiments}

\subsection{Feedback Adaptation Setup}
\label{sec:feedback_adaptation_setup}

As described in~\autoref{sec:problem}, feedback adaptation evaluates how a RAG system responds
to newly introduced feedback at a specific point in time within an otherwise continuous interaction stream.
We operationalize this setting by examining system behavior immediately before and after feedback injection at time $t$.

At time $t$, the system is assumed to have accumulated feedback from prior interactions.
To instantiate this pre-existing state in a controlled and reproducible manner,
we populate the feedback memory with synthetic feedback generated as described in~\autoref{sec:prompt_qa_generation}.
This pre-$t$ feedback serves only to stabilize retrieval and memory behavior
and does not provide task-specific supervision relevant to the evaluation queries at time $t$.

To assess adaptation, we introduce expert feedback at time $t$.
The injected feedback consists of paraphrased query–answer–context instances
that share the underlying intent with the evaluation queries while exhibiting minimal lexical overlap.
This design deliberately emphasizes semantic alignment over surface-level matching,
thereby testing the system’s ability to bridge the semantic gap inherent in retrieval.
By construction, this protocol avoids data leakage and prevents trivial keyword-based retrieval.

\paragraph{Snapshot Evaluation.}
Although feedback adaptation conceptually describes an online process,
we employ a snapshot-based evaluation protocol to isolate the marginal effect of feedback injection.
By comparing system behavior immediately before and after time $t$,
this approach controls for confounding factors such as cumulative memory drift
or changes in the query distribution.
Snapshot evaluation enables precise measurement of \textit{correction lag}
and \textit{post-feedback performance} on a well-defined set of intent-consistent but lexically distinct queries,
providing a focused test of feedback adaptation under the proposed metrics.

\setlength{\tabcolsep}{3pt}

\begin{table*}[t]
  \centering
  \small
  \begin{tabular}{lccccc}
    \toprule
    Method                                          &   Step      &         NQ      &      TriviaQA     &   HotpotQA     &    Average   \\ \cmidrule(lr){1-2} \cmidrule(lr){3-5} \cmidrule(lr){6-6} 
    Standard RAG                                    &    -     &        28.7     &        67.1       &      28.5      &     42.1 \\
    Self-RAG~\cite{asaiself}                        &    -     &        36.4     &        38.2       &      29.6      &     34.7 \\
    Auto-RAG~\cite{yu2024auto}                      &    -     &        37.9     &        60.9       &      44.9      &     47.9 \\
    ChatQA-1.5~\cite{liu2024chatqa}                 &    -     &        42.4     &        61.0       &      44.6      &     49.3 \\ \cmidrule(lr){1-6}
    Golden-only                                     &  $<t$ &        35.3     &        75.5       &      43.7      &     51.5 \\
    RAFT~\cite{zhang2024raft}                       &  $<t$ &        36.4     &        76.5       &      44.6      &     52.5 \\
\rowcolor{cyan!20}\myalg                            &  $<t$ &       36.2      &       76.8        &      44.8       &    52.6 \\ \cmidrule(lr){1-6}
    Golden-only                                     &  $t$ &        \makebox[1.4cm][r]{40.2}~{\textcolor{green!75!black}{(+4.9)}}     &       \makebox[1.4cm][r]{80.0}~{\textcolor{green!75!black}{(+4.5)}}        &      \makebox[1.4cm][r]{48.1}~{\textcolor{green!75!black}{(+4.4)}}      &     \makebox[1.4cm][r]{56.1}~{\textcolor{green!75!black}{(+4.6)}}\\
    RAFT~\cite{zhang2024raft}                       &  $t$ &        \makebox[1.4cm][r]{41.9}~{\textcolor{green!75!black}{(+5.5)}}     &       \makebox[1.4cm][r]{80.5}~{\textcolor{green!75!black}{(+4.0)}}        &      \makebox[1.4cm][r]{49.4}~{\textcolor{green!75!black}{(+4.8)}}      &     \makebox[1.4cm][r]{57.3}~{\textcolor{green!75!black}{(+4.8)}} \\
    \rowcolor{cyan!20}\myalg                        &  $t$ &       \makebox[1.55cm][r]{49.8}~{\textcolor{green!75!black}{(+13.6)}}     &   \makebox[1.4cm][r]{83.9}~{\textcolor{green!75!black}{(+7.1)}}           &      \makebox[1.4cm][r]{53.2}~{\textcolor{green!75!black}{(+8.4)}}      &     \makebox[1.4cm][r]{62.3}~{\textcolor{green!75!black}{(+9.7)}} \\ 
    % Golden-only                                     &  Oracle  &       55.1      &        82.7       &      58.7      &     65.5 \\
    % RAFT                                            &  Oracle  &       57.2      &        83.4       &      60.2      &     66.9\\
    % \rowcolor{cyan!30}\myalg                                          &  Oracle  &        76.6     &        96.5       &      84.5      &     85.9  \\   
    \bottomrule
    
    \end{tabular}
    % \vspace{-0.5em}
  \caption{\textbf{Performance comparison under feedback adaptation across three QA benchmarks.}We report performance before feedback injection (< t) and after incorporating expert feedback at time t.
{\textcolor{green!75!black}{Green numbers}} indicate post-feedback performance gains, measuring how effectively corrective feedback
generalizes to semantically related queries.
Results are reported on Natural Questions (NQ), TriviaQA, and HotpotQA using \llama-3 8B as the generator and bge-m3 as the retriever.
% We evaluate models both before feedback injection (\(< t\)) and after incorporating expert feedback at time \(t\).
% {\textcolor{green!75!black}{Green numbers}} indicate post-feedback performance gain, measured as the performance improvement after feedback is integrated.
% Results are reported on Natural Questions (NQ), TriviaQA, and HotpotQA using \llama-3 8B as the generator and bge-m3 as the retriever.
  }
  \label{tab:main}
  % \vspace{-1em}
\end{table*}

\subsection{Experimental Setup}
\label{subsec:setup}
\myparagraph{Datasets.}
We evaluate feedback adaptation performance on three standard open-domain QA benchmarks: 
Natural Questions (NQ)~\cite{kwiatkowski2019natural}, TriviaQA~\cite{joshi2017triviaqa}, and HotpotQA~\cite{yang2018hotpotqa}.  
Following common practice, we report Exact Match (EM) for NQ and TriviaQA, and F1 for HotpotQA.  
Unless otherwise noted, all methods retrieve from the full Wikipedia corpus associated with each dataset. For more details, see~\autoref{sec:implementation}.

\myparagraph{Feedback adaptation metrics.}
As described in~\autoref{subsec:correction_lag} and~\autoref{subsec:post_feedback}, we evaluate feedback adaptation using two distinct metrics.
\emph{Correction lag} is measured as the wall-clock time from when feedback is provided until the system is ready to process subsequent evaluation queries using the updated state.
\emph{Post-feedback performance} is measured as the system’s performance (\ie EM or F1 score) at step $t$ after the update, evaluated on queries that require the same correction implied by the feedback.

\myparagraph{Implementation details.}
We compare \myalg against (\emph{i}) non-feedback RAG enhancement methods such as Self-RAG~\cite{asaiself}, Auto-RAG~\cite{yu2024auto}, and ChatQA-1.5~\cite{liu2024chatqa}, and (\emph{ii}) a feedback-driven adaptation baseline, RAFT~\cite{zhang2024raft} (See~\autoref{app:train_conf} for training details).
For fairness, all systems use the same \llama-3 8B generator~\cite{dubey2024llama}.
For retrieval, we employ bge-m3~\cite{chen2024bge} as a retriever to generate embeddings for both user intents and evidence contexts within a shared vector space.
Additionally, we populate the feedback memory with synthetic instances scaling to the full size of the retrieval corpus (\eg $\sim$150K for TriviaQA) to simulate a realistic deployed state.
The weighting parameter $\lambda$ in~\autoref{eq:patchrag_sim} is fixed to $0.5$ across all experiments.
All experiments are conducted on two NVIDIA A5000 GPUs and an Intel Xeon Gold 6342 CPU.

\begin{figure}[t]
    \centering
    \includegraphics[width=0.95\linewidth]{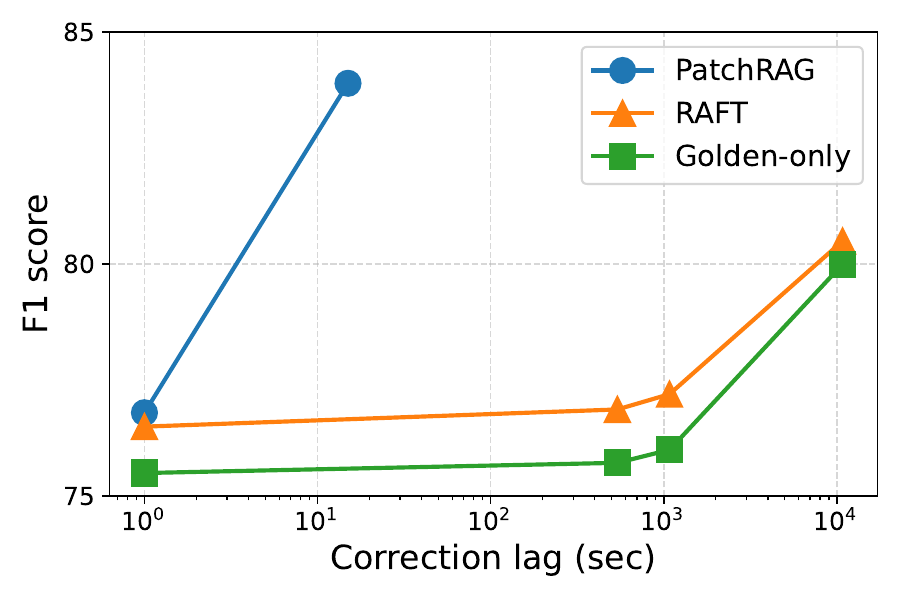}
    % \vspace{-0.5cm}
    \caption{\textbf{Correction lag vs. Post-feedback Performance (F1) on TriviaQA.}
    This figure illustrates a structural trade-off inherent to training-based feedback incorporation, which inference-time methods mitigate.
    To investigate the trade-off between latency and performance, we evaluated training-based baselines (RAFT, Golden-only) under varying compute budgets by reducing update steps from full fine-tuning down to $\sim$5\% of steps (See~\autoref{sec:ablation}).
    }
    \label{fig:correction_lag}
    % \vspace{-1em}
\end{figure}

\subsection{Experimental Results}
We evaluate \myalg under a feedback adaptation setting to assess how efficiently it incorporates corrective feedback.
Following the metrics introduced in~\autoref{subsec:correction_lag} and~\autoref{subsec:post_feedback}, we analyze system behavior along two dimensions:
(\emph{i}) correction lag and
(\emph{ii}) post-feedback performance.

\myparagraph{\myalg improves performance before expert feedback is introduced.}
As shown in~\autoref{tab:main}, incorporating pre-$t$ feedback stabilizes retrieval behavior, providing a stronger initial state for subsequent feedback adaptation. Even before expert feedback is injected, this stabilization leads to consistent performance gains over models that do not leverage feedback. In particular, the dual intent-context retrieval mechanism reduces early-stage retrieval mismatch by better aligning retrieved evidence with the underlying query intent.

\myparagraph{\myalg achieves the best post-feedback performance.}
We next evaluate post-feedback performance, indicated by the green numbers in~\autoref{tab:main}, which measures how reliably a system produces corrected behavior after expert feedback is incorporated at step~$t$.
In this setting, \myalg achieves the highest absolute accuracy across all three QA benchmarks, with an average score of 62.3, surpassing training-based baselines such as Golden-only and RAFT.
Furthermore, \myalg exhibits the largest post-feedback performance gain, with an average improvement of 9.7.
This result validates post-feedback performance as a distinct evaluation axis, revealing differences that are obscured by standard accuracy metrics.
% This indicates that the corrective signals generalize effectively to semantically related queries, leading to stronger behavioral reliability than competing approaches.

\myparagraph{\myalg has the lowest correction lag while achieving the highest post-feedback performance.}
We finally analyze correction lag, defined as the time required for a system to reflect newly provided feedback in its predictions.
As illustrated in~\autoref{fig:correction_lag}, training-based methods (\eg RAFT) exhibit a fundamental trade-off: deliberately reducing the number of update steps to shorten correction lag results in a sharp degradation of post-feedback performance.
Even under these strictly constrained training budgets, training-based approaches fail to match the immediate responsiveness required for feedback adaptation.
Conversely, \myalg avoids this trade-off by incorporating feedback at inference time, achieving low correction lag without sacrificing post-feedback performance.
% circumvents this trade-off entirely, achieving low-correction lag updates while surpassing the post-feedback performance of fully fine-tuned baselines.

\subsection{Robustness to Imperfect Feedback}
\label{subsec:stress_feedback_quality}

To evaluate whether \myalg remains effective under more realistic feedback conditions, we further stress-test the quality of corrective feedback. In addition to the clean setting used in the main experiments, we consider five imperfect-feedback variants: replacing gold evidence with retrieved top-1 evidence, corrupting the answer field in a fraction of feedback instances (\textbf{Noise-X\%}), omitting the answer entirely (\textbf{Blank}), providing the answer only implicitly in a verbose explanation (\textbf{Vague}), and storing both correct and incorrect feedback for the same intent (\textbf{Conflict}).

\begin{table}[t]
\centering
\small
\begin{tabular}{llccc}
\toprule
\textbf{Answer} & \textbf{Evidence} & \textbf{NQ} & \textbf{TriviaQA} & \textbf{HotpotQA} \\
\midrule
G.T.         & G.T.   & 49.8 & 83.9 & 53.2 \\
G.T.         & Top-1  & 49.2 & 81.6 & 52.5 \\
Noise (25\%) & Top-1  & 48.7 & 81.0 & 49.9 \\
Noise (50\%) & Top-1  & 44.1 & 78.0 & 42.6 \\
Noise (75\%) & Top-1  & 37.2 & 74.8 & 32.5 \\
Blank        & Top-1  & 33.2 & 77.2 & 37.1 \\
Vague (Long) & Top-1  & 34.7 & 77.5 & 39.8 \\
Conflict     & Top-1  & 43.9 & 78.2 & 42.4 \\
\bottomrule
\end{tabular}
\caption{\textbf{Stress-test results under imperfect feedback conditions.} We report EM on Natural Questions (NQ) and TriviaQA, and F1 on HotpotQA.}
\label{tab:stress_feedback_quality}
\end{table}

\setlength{\tabcolsep}{5pt}
\begin{table}[t]
    \centering
    \small 
    % \begin{adjustbox}{width=0.98\linewidth}
    \begin{tabular}{@{}lccc@{}}
    \toprule
    Method                                          &       NQ       &      TriviaQA       &      HotpotQA             \\\cmidrule(lr){1-4}
    Standard RAG                                    &      28.7      &      67.1           &       28.5                \\
    $~~+$ Retr.~\autoref{eq:patchrag_sim}             &      45.6      &      77.8           &       47.6               \\
    \textbf{$~~+$ Add feedback (ours)}                           &      \textbf{49.8}      &      \textbf{83.9}           &       \textbf{53.2}          \\
    \bottomrule
    \end{tabular}
    % \end{adjustbox}
    
    \caption{\textbf{Ablation study under feedback adaptation.}
We incrementally introduce feedback-related components into a standard RAG system
and evaluate post-feedback performance.
Replacing standard retrieval scoring with the dual-similarity metric improves reliability
by enabling intent-aware feedback retrieval.
Adding feedback exemplars further strengthens post-feedback behavior,
highlighting the necessity of exemplar-based conditioning for effective feedback adaptation.
  } 
  \label{tab:ablation}
  % \vspace{-3.0em}
\end{table}

As shown in \autoref{tab:stress_feedback_quality}, replacing gold evidence with automatically retrieved top-1 evidence causes only a small drop (49.8$\rightarrow$49.2 on NQ, 83.9$\rightarrow$81.6 on TriviaQA, and 53.2$\rightarrow$52.5 on HotpotQA), indicating that \myalg does not rely heavily on explicitly curated evidence and remains effective when feedback context must be retrieved automatically.

\myalg also degrades gradually as the corrective signal becomes noisier. Under answer poisoning, performance decreases monotonically from Noise 25\% to Noise 75\%, rather than collapsing abruptly, suggesting that the method tracks the strength of the available corrective signal and retains useful adaptation under moderate corruption.

The most challenging cases arise when the corrective signal is weak or inconsistent. Both \textbf{Blank} and \textbf{Vague} feedback substantially reduce performance, showing that correction requires an explicit and retrievable answer signal rather than additional text alone. The \textbf{Conflict} setting also degrades performance, although it remains above the most severely corrupted noise condition, suggesting that retrieval-based adaptation can partially tolerate contradictory feedback.

Overall, our stress tests broaden feedback adaptation evaluation by explicitly covering practically relevant imperfect-feedback conditions encountered in real-world deployment.

\subsection{Further Analyses}
\label{sec:further}
We conduct further analyses to better understand the structural requirements of feedback adaptation.
Rather than comparing methods in isolation, these analyses examine
(\emph{i}) why standard retrieval paradigms fail to propagate feedback,
(\emph{ii}) which components are necessary for reliable adaptation,
(\emph{iii}) how grounding and generalization trade off during adaptation, and
(\emph{iv}) whether these behaviors persist across different retrieval backbones.
% Together, these analyses provide insight into the conditions under which corrective feedback can be reliably internalized and generalized.

\begin{figure}[t] 
    \centering
    \includegraphics[width=0.9\linewidth]{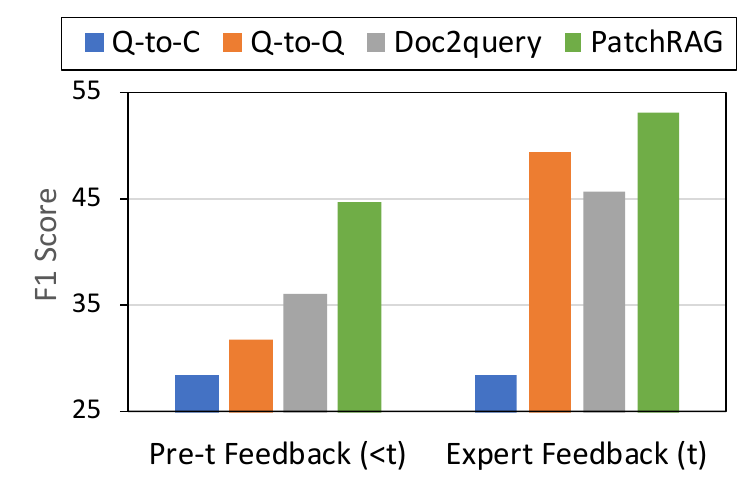}
    \vspace{-0.5em}
    % \caption{\textbf{Ablation study of query-based retrieval strategies on HotpotQA.}
    % We compare Q-to-Q (query-to-query matching) and Q-to-C (query-to-context matching; \ie the standard RAG retrieval formulation), and Doc2query~\cite{nogueira2019document} with our proposed method under different feedback settings.
    % }
     \caption{\textbf{Ablation of query-based retrieval strategies under feedback adaptation on HotpotQA.} We compare query-to-context (Q-to-C; standard RAG), query-to-query (Q-to-Q), and Doc2query~\cite{nogueira2019document} with an intent-context retrieval strategy before and after expert feedback injection. The results illustrate how intent-level retrieval becomes critical for propagating corrective feedback under semantic variation.}
    \label{fig:retrieval}
    % \vspace{-1em}
\end{figure}

\setlength{\tabcolsep}{6pt}
\begin{table*}[t]
  \centering
  \small
  % \begin{adjustbox}{width=0.98\linewidth}
  \begin{tabular}{@{}lccccc@{}}
    \toprule
    $\mathcal{R}$                                      &       Step                   &     NQ      &   TriviaQA  &   HotpotQA     &    Average   \\ \cmidrule(lr){1-2} \cmidrule(lr){3-5} \cmidrule(lr){6-6}
                    \multirow{2}{*}{BM25}              &     $<t$                &    26.0     &    75.6     &    30.1        &   43.9   \\
                                                       &     $t$                 &    \makebox[1.4cm][r]{27.0}~{\textcolor{green!75!black}{(+1.0)}}     &    \makebox[1.4cm][r]{77.8}~{\textcolor{green!75!black}{(+2.2)}}     &    \makebox[1.4cm][r]{30.8}~{\textcolor{green!75!black}{(+0.7)}}        &   \makebox[1.4cm][r]{45.2}~{\textcolor{green!75!black}{(+1.3)}}   \\ \cmidrule(lr){1-6}
                                                       % &     Oracle    &    27.1     &    90.1     &    31.2        &   49.5   \\ \cmidrule(lr){1-6}
    
                    \multirow{2}{*}{DPR}               &     $<t$                &    36.9     &    74.7     &    43.8        &   51.8   \\
                                                       &     $t$   &    \makebox[1.55cm][r]{50.0}~{\textcolor{green!75!black}{(+13.1)}}     &    \makebox[1.4cm][r]{81.6}~{\textcolor{green!75!black}{(+6.9)}}     &    \makebox[1.4cm][r]{52.7}~{\textcolor{green!75!black}{(+8.9)}}        &   \makebox[1.4cm][r]{61.4}~{\textcolor{green!75!black}{(+9.6)}}   \\ \cmidrule(lr){1-6}
                                                       % &     Oracle    &     77.9     &        93.2       &      81.3      &     84.1   \\ \cmidrule(lr){1-6}

                    \multirow{2}{*}{bge-m3}            &     $<t$               &    36.2     &    76.8     &    44.8        &   52.6   \\
                                                       &     $t$   &    \makebox[1.55cm][r]{49.8}~{\textcolor{green!75!black}{(+13.6)}}     &    \makebox[1.4cm][r]{83.9}~{\textcolor{green!75!black}{(+7.1)}}     &    \makebox[1.4cm][r]{53.2}~{\textcolor{green!75!black}{(+8.4)}}      &   \makebox[1.4cm][r]{62.3}~{\textcolor{green!75!black}{(+9.7)}}     \\
                                                       % &     Oracle   &    76.6     &    96.5     &    84.5        &   85.9   \\ 
    \bottomrule
    
    \end{tabular}
    % \end{adjustbox}
    % \vspace{-0.5em}
    % \caption{\textbf{\myalg performance across different retrievers under feedback adaptation.}
    % We report results before feedback injection (\(< t\)) and after incorporating expert feedback at time \(t\).
    % {\textcolor{green!75!black}{Green numbers}} indicate post-feedback performance gain, measured as performance improvement after feedback is integrated.
    % Across BM25, DPR, and bge-m3, \myalg consistently benefits from feedback, with stronger dense retrievers yielding larger gains, highlighting the role of embedding quality in propagating corrections to semantically related queries.
    % }
    \caption{\textbf{Performance across different retrieval backbones under feedback adaptation.} We report results before (< t) and after incorporating expert feedback at time t. {\textcolor{green!75!black}{Green numbers}} indicate post-feedback performance gains. Across sparse (BM25) and dense (DPR, bge-m3) retrievers, incorporating feedback consistently improves post-feedback performance, with stronger embedding spaces enabling more reliable propagation to semantically related queries.}
  \label{tab:retriever}
  % \vspace{-0.5em}
\end{table*}

\myparagraph{Why standard retrieval fails under feedback adaptation.}
We first analyze why conventional retrieval strategies struggle to propagate corrective feedback.
\autoref{fig:retrieval} compares four query-based retrieval approaches on HotpotQA:
Q-to-C (standard RAG retrieval), Q-to-Q matching, Doc2query~\cite{nogueira2019document},
and the proposed dual intent--context retriever.
Standard Q-to-C retrieval exhibits the weakest performance and shows almost no improvement after expert feedback is introduced.
This behavior highlights a fundamental limitation of content-only retrieval.
Q-to-Q matching substantially improves performance after feedback injection with 21.0 points, indicating that intent-level similarity becomes critical once corrections are available. Doc2query improves pre-$t$ performance by expanding lexical coverage,
but yields limited gains (+7.6 points) after expert feedback, reflecting the constraints of operating within a single embedding space.

In contrast, combining both intent and context signals enables both grounding
and semantic generalization, allowing feedback to propagate reliably to related queries.
These results illustrate why standard retrieval paradigms are insufficient
for feedback adaptation under semantic variation.

\myparagraph{Component analysis.}
We examine which components are required to support reliable post-feedback behavior.~\autoref{tab:ablation} presents an ablation study of key system components for feedback adaptation.
Replacing standard retrieval scoring with the dual-similarity formulation
substantially improves post-feedback performance,
demonstrating that intent-aware retrieval is necessary to surface relevant corrective feedback.
In addition, explicitly incorporating retrieved feedback into the generation prompt
further improves reliability after feedback injection.
This indicates that feedback adaptation requires not only retrieving corrections,
but also conditioning generation on explicit corrective exemplars.

Together, these results suggest that reliable feedback adaptation depends on
both intent-level retrieval and feedback-conditioned generation,
rather than on either component in isolation.

\begin{figure}[t] 
    \centering
    \includegraphics[width=0.95\linewidth]{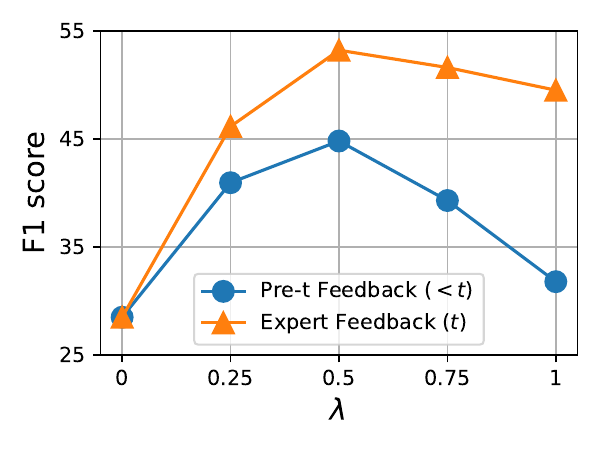}
    \vspace{-1.0em}
    % \caption{\textbf{Impact of the balancing parameter~$\lambda$ on HotpotQA.} The best performance is achieved at $\lambda=0.5$, while larger values lead to a gradual decline.
    % }
    \caption{\textbf{Impact of the balancing parameter $\lambda$ on feedback adaptation behavior on HotpotQA.} We report performance before and after expert feedback injection. Varying $\lambda$ highlights the balance between content grounding and semantic generalization, which influences the reliability of post-feedback behavior.}
    \label{fig:hyperparam}
    % \vspace{-1em}
\end{figure}

\myparagraph{Intent-context trade-off in feedback adaptation.}
We analyze how feedback adaptation behavior varies along the intent-context axis by examining the effect of the balancing parameter~$\lambda$
in the intent-context retrieval mechanism.
\autoref{fig:hyperparam} reports performance on HotpotQA before and after expert feedback injection.

When only pre-$t$ feedback is available, performance peaks at intermediate $\lambda$ values, indicating that excessive reliance on either content grounding or intent generalization can degrade retrieval reliability. After expert feedback is incorporated at step~$t$, performance becomes less sensitive to larger $\lambda$ values, reflecting improved alignment once explicit corrections are available.

These trends highlight a trade-off in feedback adaptation: effective correction requires balancing semantic generalization with content grounding. We adopt $\lambda=0.5$ as a stable operating point that consistently balances these requirements across feedback settings.

\myparagraph{Robustness across retrieval backbones.}
Finally, we evaluate whether feedback adaptation behavior depends on a specific retrieval backbone.~\autoref{tab:retriever} reports results for BM25, DPR, and bge-m3 under different feedback availability settings.

Across all retrievers, incorporating expert feedback at step~$t$ consistently improves post-feedback performance, indicating that feedback adaptation generalizes across heterogeneous retrieval settings. Sparse retrieval with BM25 exhibits smaller gains due to limited semantic matching, which constrains intent-level generalization. Dense retrievers such as DPR and bge-m3 yield larger improvements, reflecting higher-quality embedding spaces that better support feedback retrieval.
\section{Related Works}
\myparagraph{Evaluation of RAG and knowledge adaptation.}
RAG systems are commonly evaluated using static open-domain QA benchmarks such as TriviaQA~\cite{joshi2017triviaqa}, Natural Questions~\cite{kwiatkowski2019natural}, and HotpotQA~\cite{yang2018hotpotqa}.
While these standard benchmarks rigorously assess retrieval and reasoning on fixed knowledge snapshots, they fail to measure how systems adapt to errors over time.
To address model updates, recent research has expanded into knowledge editing~\cite{meng2022locating, mengmass} and memory-centric RAG~\cite{jimenez2024hipporag, packer2023memgpt}, focusing on mechanisms to construct or edit memory structures.
However, a critical gap remains in evaluation: neither static QA benchmarks nor editing-specific metrics quantify the operational dynamics of feedback adaptation---the correction lag and post-feedback performance in a deployed stream.
In this work, we address this limitation by formalizing \textit{feedback adaptation} to explicitly measure these temporal trade-offs.

% \myparagraph{RAG evaluation benchmarks.}
% RAG systems are commonly evaluated using static open-domain QA benchmarks. Single-hop retrieval and generation performance is typically measured with datasets such as TriviaQA~\cite{joshi2017triviaqa} and Natural Questions~\cite{kwiatkowski2019natural}, while multi-hop or implicit reasoning capabilities are assessed through HotpotQA~\cite{yang2018hotpotqa} and MuSiQue~\cite{trivedi2022musique}. Broader benchmark suites including KILT~\cite{petroni2021kilt} and BEIR~\cite{thakur2021beir} integrate diverse knowledge-intensive tasks and retrieval settings into unified evaluation frameworks. These benchmarks, however, evaluate models on fixed snapshots of knowledge, leaving the question of how RAG systems adapt to corrective feedback over time largely understudied.

\myparagraph{RAG adaptation for specific domains.}
Recent studies view domain-specific RAG as a training-time adaptation problem that jointly optimizes retrieval and generation components. RAG-end2end~\cite{siriwardhana2023improving} demonstrates that end-to-end training improves robustness to domain shifts and noisy retrieval compared to modular pipelines. To further strengthen grounding, RAFT~\cite{zhang2024raft} trains language models with both relevant and distracting documents, promoting selective evidence usage and citation-aware reasoning in open-book settings. Complementary to model-centric approaches, data-centric methods~\cite{mao2024rag, xu2025simrag} exploit unlabeled in-domain corpora to synthesize contrastive QA pairs, enabling self-supervised RAG adaptation without manual annotations. Beyond unstructured text retrieval, recent works~\cite{barron2024domain, sharma2025og} integrates structured knowledge sources to retrieve compact and verifiable contexts, effectively reducing hallucinations in technical domains. However, they are not designed for direct feedback-driven adaptation and rely on training-time updates, leading to substantial correction lag.

\myparagraph{RAG Enhancement.}
Efforts to enhance RAG span both training-time and inference-time methodologies, each aiming to improve evidence utilization and robustness. Train-based approaches optimize the retriever, generator, or both: RA-DIT~\cite{linra} and InstructRetro~\cite{wanginstructretro} fine-tune LLMs with retrieval-augmented instructions; RankRAG~\cite{yu2024rankrag}, Chain-of-Note~\cite{yu2023chain}, ChatQA~\cite{liu2024chatqa}, and Self-RAG~\cite{asaiself} strengthen generation robustness to noisy or incomplete retrieval; and preference-based methods such as RPO~\cite{yan2025rpo} promote faithful grounding. However, these techniques rely on parameter updates, making adaptation to new user feedback dependent on additional fine-tuning. In contrast, train-free strategies improve RAG at inference time by optimizing queries and retrieved contexts, via hypothetical document generation~\cite{gao2023precise} and multi-query fusion~\cite{rackauckas2024rag}, or by employing LLM-based listwise reranking~\cite{sun2023chatgpt, sachan2022improving}. More recent directions leverage inference-time scaling and structural signals, with graph-based methods like GraphRAG~\cite{edge2024local} and LightRAG~\cite{guo2024lightrag}, and agentic or reasoning-driven designs such as MemoRAG~\cite{qian2025memorag} and TSSS~\cite{bang2025think}. Overall, while these methods substantially improve RAG effectiveness, they leave the direct incorporation of user feedback understudied.

\myparagraph{FAQ-based retrieval.}
To mitigate the semantic gap between user queries and declarative documents, several approaches propose transforming the retrieval task into a query-to-query matching problem, effectively constructing a synthetic FAQ index. Doc2Query~\cite{nogueira2019document} pioneered this direction by expanding documents with predicted queries. With the advent of LLMs, methods like InPars~\cite{bonifacio2022inpars} have advanced this paradigm by generating high-quality synthetic QA pairs via few-shot prompting. 
\section{Conclusion}
We introduced \textit{feedback adaptation} as a problem setting for evaluating RAG systems under interactive feedback. We proposed two complementary evaluation axes, \textit{correction lag} and \textit{post-feedback performance}, to capture adaptation behavior that is overlooked by standard accuracy-based evaluation. Through empirical analysis, we showed that training-based approaches exhibit a trade-off between delayed correction and reliable adaptation. We further presented \myalg as a minimal inference-time instantiation that illustrates the practical implications of feedback adaptation.
% We formalize \emph{feedback adaptation} as a critical evaluation paradigm, identifying \emph{correction lag} as the central barrier to responsive RAG systems.
% While training-based approaches fundamentally struggle with update latency, our proposed \myalg overcomes this limitation through a lightweight, inference-time dual-retrieval mechanism.
% By effectively resolving the trade-off between speed and reliability, we aim to establish feedback adaptation as a key research frontier, encouraging the development of diverse future solutions for truly dynamic RAG deployments.

\section{Limitations}
This work focuses on feedback adaptation in retrieval-augmented generation, specifically on how corrective feedback propagates through retrieval and generation once it is introduced. Accordingly, we evaluate adaptation under a controlled setting that enables precise measurement of correction lag and post-feedback behavior. While real-world feedback may be more heterogeneous, noisy, or interactive, these complexities primarily affect feedback acquisition rather than the propagation mechanisms studied in this work. Extending the framework to richer human-in-the-loop feedback settings is an important direction for future research, but does not undermine the validity of the feedback adaptation behaviors evaluated here.

Our approach performs adaptation through non-parametric memory updates and inference-time conditioning, enabling immediate behavioral adjustment without retraining. This design choice intentionally targets deployment scenarios where rapid response to feedback is critical. However, as feedback accumulates over time, unresolved or conflicting feedback may require higher-level consolidation or reconciliation mechanisms. Such long-horizon consistency management is complementary to the feedback adaptation framework introduced in this work and remains an open challenge.

\section{Potential Risk}
Feedback adaptation in retrieval-augmented generation relies on storing and retrieving external feedback and documents, which may include sensitive or private information.
While \myalg focuses on the mechanisms of feedback propagation rather than data collection, improper handling of feedback memory could raise privacy concerns in real-world deployments. As with other RAG-based systems, practical applications should ensure that external databases and feedback stores comply with data protection and privacy regulations, and that sensitive information is handled appropriately.

\bibliography{custom}

\newpage
\appendix
\appendix 
\supptitle

\section{Prompt Template for RAG}
\label{sec:prompt_rag}

Prompt engineering with retrieved chunks or Q\&As is another important factor to improve performance. For fair comparion between \myalg and standard RAG, we unified the prompt template. 

\myparagraph{Justification for Q\&A inclusion.} 
The inclusion of explicit Q\&A pairs in the prompt is a deliberate design choice to simulate a feedback adaptation scenario where the system must prioritize and adhere to specific expert corrections. Crucially, the answer components provided in-context are synthetic responses generated by an auxiliary model (GPT-4) to mimic human feedback(See~\autoref{sec:prompt_qa_generation}), and are independent of the ground-truth labels used for final evaluation. Consequently, this setup does not constitute data leakage; rather, it rigorously evaluates the system's ability to generalize correction instructions to lexically distinct queries through semantic retrieval, measuring behavioral alignment rather than trivial answer copying. 

\begin{tcolorbox}[fonttitle=\small\bfseries,
fontupper=\scriptsize\sffamily,
fontlower=\fon{put},
enhanced,
left=2pt, right=2pt, top=2pt, bottom=2pt,
title=Prompt template for \myalg{}]
\begin{lstlisting}[]
Question: {QUESTION #1}
Answer: {ANSWER #1} 
Question: {QUESTION #2}
Answer: {ANSWER #2}
...
Question: {QUESTION #5}
Answer: {ANSWER #5}
Context1: {CONTEXT #1}
Context2: {CONTEXT #2}
...
ContextN: {CONTEXT #N}
Please answer the below question based on given 
above question and answer pairs and contexts. 
Note that you should generate the response only 
for answering the question within a few words. 
Do not contain extra comments.
Question: {TARGET QUESTION}
\end{lstlisting}
\end{tcolorbox}

\begin{tcolorbox}[fonttitle=\small\bfseries,
fontupper=\scriptsize\sffamily,
fontlower=\fon{put},
enhanced,
left=2pt, right=2pt, top=2pt, bottom=2pt,
title=Prompt template for standard RAG]
\begin{lstlisting}[]
Context1: {CONTEXT #1}
Context2: {CONTEXT #2}
...
Context5: {CONTEXT #5}
Please answer the below question based on given 
above contexts. Note that you should generate 
the response only for answering the question 
within a few words. Do not contain extra comments.
Question: {TARGET QUESTION}
\end{lstlisting}
\end{tcolorbox}

\section{Feedback Preparation}
\label{sec:prompt_qa_generation}
\subsection{Pre-$t$ Feedback Generation}
As described in~\autoref{subsec:setup}, the feedback available before step $t$ (\ie pre-$t$ feedback) is synthetically generated using \llama-3~8B. 
For each document chunk in the corpus, we prompt \llama-3~8B to generate corresponding question--answer pairs using the template shown below. 
The resulting outputs are parsed into structured Q\&A pairs and stored as feedback instances in the memory.

\begin{tcolorbox}[fonttitle=\small\bfseries,
fontupper=\scriptsize\sffamily,
fontlower=\fon{put},
enhanced,
left=2pt, right=2pt, top=2pt, bottom=2pt,
title=Prompt template for Q\&A Generation from LLM]
\begin{lstlisting}[]
please generate question and answer pairs to check 
the understandability of following context. 
Format should be Q: question A: answer.
{CONTEXT}
\end{lstlisting}
\end{tcolorbox}

\subsection{Expert Feedback at Step $t$}
Expert feedback at step $t$ is constructed by first paraphrasing each evaluation query using GPT-4. 
The paraphrased query is then provided to GPT-4 together with the corresponding evidence context, and the resulting response is used as the corrected answer in the feedback instance. 
Each feedback entry thus consists of a paraphrased query, a generated answer, and its supporting context, simulating an expert correction with the same underlying intent as the evaluation query.

To ensure a clean evaluation protocol, the original test queries are not directly stored in the feedback memory. 
The paraphrased queries differ in surface form from the evaluation queries and are used solely to provide intent-aligned but lexically distinct feedback. 
This design allows us to assess whether corrections generalize to semantically related inputs rather than relying on direct exposure to the test instances.

\section{Implementation Details}
\label{sec:implementation}
\subsection{Datasets}
\textbf{TriviaQA}~\cite{joshi2017triviaqa}. It offers a complex question answering dataset featuring 950,000 pairs of questions and answers sourced from over 662,000 documents on Wikipedia and the internet. Unlike simpler QA benchmarks like SQuAD, TriviaQA presents a tougher challenge because answers often require more than just predicting a text span from lengthy contexts. The dataset includes both subsets that are verified by humans and those created by machines, and we only utilized verified datasets in this experiment.

\noindent \textbf{Natural Questions}~\cite{kwiatkowski2019natural}. It is designed for training question answering systems, featuring over 300,000 training, nearly 8,000 development, and approximately 8,000 test examples, each pairing a Google search query with a relevant Wikipedia article. These articles include marked sections that provide a detailed response to the query, as well as shorter snippets that directly answer the question, although some annotations may be absent, indicating no answer is available. Additionally, a small fraction of the dataset, about 1\%, contains binary "yes" or "no" answers instead of detailed excerpts.

\noindent \textbf{HotpotQA}~\cite{yang2018hotpotqa}. It is a dataset featuring roughly 113,000 questions based on English Wikipedia, designed to need the lead sections from two articles for answers. Accompanying each question are the key paragraphs and a selection of sentences marked by crowdworkers as essential facts for responding. However, there are several examples that are missing some key paragraphs, and it causes the performance drop in RAG. 

\setlength{\tabcolsep}{16pt}
\begin{table}[t]
  \centering
  \scriptsize
  % \resizebox{1.0\linewidth}{!}{
  \begin{tabular}{@{}p{3.5cm}r@{}}
    \toprule
    Parameters         &         Value              \\\cmidrule(lr){1-2}
    Rank               &        8                  \\ 
    Alpha              &        16                  \\
    Train on Inputs    &        True                \\ 
    LoRA Modules       &        Q, K, V, O          \\
    Dropout            &         0.05               \\

    Batch Size         &        128                  \\
    Loss Function      &        Cross Entropy Loss   \\
    Learning Rate      &        3e-4                 \\
    Scheduler          &        Cosine Annealing     \\ 
    Optimizer          &        AdamW                \\
    Epochs             &         1                  \\
    \bottomrule
    \end{tabular}
    % }
  \caption{\textbf{Training hyperparameters.}
  } 
  \vspace{-1em}
  \label{tab:parameters}
\end{table}

\setlength{\tabcolsep}{6pt}

\subsection{Evaluation Metrics}
\textbf{EM}. Exact Match is a strict metric that measures the percentage of predictions that match the ground truth exactly. It is often used in the context of machine comprehension and question answering tasks where the goal is to produce an exact answer. For a given dataset, the Exact Match score is calculated by dividing the number of predictions that are exactly the same as the true answers by the total number of predictions made.

\noindent \textbf{F1}. The F1 score is a more nuanced metric that considers both precision and recall. It is the harmonic mean of precision and recall, providing a balance between the two.

\section{Training Configuration}
\label{app:train_conf}

To validate the superiority of our method compared to train-based RAG, RAFT and golden-only finetuning methods followed the parameters as described in \autoref{tab:parameters}. We adopted LoRA tuning~\cite{hulora} that is one of the parameter efficient finetuning methods. Despite of imbalance problem of three datasets, both RAFT and golden-only methods get higher EM as using more relevant dataset as a training set, but they have lower performance than \myalg{} that is only updating the dataset into the database. Lastly, all the experiments for training are conducted on two Nvidia A5000s.

\section{Further Analyses}
\label{sec:ablation}

\setlength{\tabcolsep}{18pt}
\begin{table}[t]
    \centering
    
    \scriptsize 
    % \begin{adjustbox}{width=0.98\linewidth}
    \begin{tabular}{@{}lccc@{}}
    \toprule
    Method                     &       Retrieval       &      Generation       &      Total     \\\cmidrule(lr){1-4}
    RAFT                       &        23.7           &        459.4          &      483.2        \\
    \myalg{}                   &        24.6           &        450.5          &      475.2        \\
    
    \bottomrule
    \end{tabular}
    % \end{adjustbox}
    
    \caption{\textbf{Inference time (milliseconds) comparison between RAFT and \myalg{}}. 
  } 
  \vspace{-1em}
  \label{tab:inference_cost}
\end{table}
\setlength{\tabcolsep}{6pt} 

\subsection{Inference Computation Cost}
To assess the inference computation cost, we measured the inference time of \myalg{} and train-based RAG (\ie RAFT) using dense retrieval with DPR. The inference time was averaged over 100 examples after a warm-up run with another 100 examples, conducted on a server equipped with two NVIDIA A5000 GPUs and an Intel Xeon Gold 6342 CPU @ 2.80GHz. As shown in~\autoref{tab:inference_cost}, while train-based RAG achieves faster retrieval (23.7 ms vs. 24.6 ms), it incurs higher generation latency (459.4 ms vs. 450.5 ms), leading to a slightly higher total inference time (483.2 ms vs. 475.2 ms). This increased generation time stems from the integration of LoRA into the base LLM in our setup. Although train-based RAG benefits from marginally faster retrieval, the overall time difference is negligible. This suggests that inference-time feedback incorporation does not impose prohibitive overhead in practice.

\setlength{\tabcolsep}{13pt}
\begin{table*}[t]
  \centering
  \scriptsize
  % \resizebox{1.0\linewidth}{!}{
  \begin{tabular}{@{}lll@{}}
    \toprule
    \multicolumn{3}{c}{\textsf{Question: What do you practice in a dojo?  Answer: martial art}} \\\cmidrule(lr){1-3}
    Method             &         Type               &         \multicolumn{1}{c}{Contents}          \\\cmidrule(lr){1-3}
    
    \multirow{5}{*}{\makecell[c]{Standard\\RAG}} & context            &        \makecell[l]{Sport Gichin Funakoshi said, "There are no contests in karate." In pre–World War II Okinawa, kumite was not part of\\ 
                                                                                            karate training. Shigeru Egami relates that, in 1940, some karateka were ousted from their dojo because they adopted\\
                                                                                            sparring after having learned it in Tokyo. Karate is divided into style organizations. These organizations sometimes\\
                                                                                            cooperate in non-style specific sport karate organizations or federations. Examples of sport organizations include\\
                                                                                            AAKF/ITKF, AOK, TKL, AKA, WKF, NWUKO, WUKF and WKC. Organizations hold competitions (tournaments)\\
                                                                                            from local to international level. Tournaments are designed to match}            \\\cmidrule(lr){2-3}
                                                 & context           &         \makecell[l]{and in some rare cases even time-limited grappling on the ground are also allowed. Free sparring is performed in a \\
                                                                                            marked or closed area. The bout runs for a fixed time (2 to 3 minutes.) The time can run continuously (iri kume) or \\
                                                                                            be stopped for referee judgment. In light contact or semi contact kumite, points are awarded based on the criteria: \\
                                                                                            good form, sporting attitude, vigorous application, awareness/zanshin, good timing and correct distance. In full \\
                                                                                            contact karate kumite, points are based on the results of the impact, rather than the formal appearance of the scoring\\
                                                                                            technique.}\\ \cmidrule(lr){1-3}
   \multirow{18}{*}{\myalg{}}          &      context               &          \makecell[l]{Karate appeared in the Soviet Union in the mid-1960s, during Nikita Khrushchev's policy of improved international \\
                                                                                            relations. The first Shotokan clubs were opened in Moscow's universities. In 1973, however, the government banned \\
                                                                                            karate—together with all other foreign \hl{martial arts}—endorsing only the Soviet \hl{martial art} of sambo. Failing \\
                                                                                            to suppress these uncontrolled groups, the USSR's Sport Committee formed the Karate Federation of USSR in December \\
                                                                                            1978. On 17 May 1984, the Soviet Karate Federation was disbanded and all karate became illegal again. In 1989, karate\\ 
                                                                                            practice became legal again, but under strict government regulations, only after the dissolution of the Soviet Union \\
                                                                                            in 1991 did independent karate schools resume functioning, and so federations were formed and national tournaments in \\
                                                                                            authentic styles began.}\\ \cmidrule(lr){2-3}
                                      &     context                  &         \makecell[l]{Shigeru Egami, Chief Instructor of Shotokan Dojo, opined that "the majority of followers of karate in overseas \\
                                                                                            countries pursue karate only for its fighting techniques...Movies and television... depict karate as a mysterious way \\
                                                                                            of fighting capable of causing death or injury with a single blow... the mass media present a pseudo art far from the \\
                                                                                            real thing." Shoshin Nagamine said, "Karate may be considered as the conflict within oneself or as a life-long marathon \\
                                                                                            which can be won only through self-discipline, hard training and one's own creative efforts.}\\ \cmidrule(lr){2-3}
                                      &     Q\&A          &       \makecell[l]{Question: When did karate practice become legal again in the Soviet Union? \\ 
                                                                               Answer: In 1989}\\\cmidrule(lr){2-3}
                                      &     Q\&A          &       \makecell[l]{Question: What do Shigeru Egami think is the main reason why people in overseas countries practice karate? \\
                                                                               Answer: They practice karate only for its fighting techniques.}\\\cmidrule(lr){2-3}
                                      &     Q\&A          &       \makecell[l]{Question: What did the JKA do for Bell's grades? \\
                                                                               Answer: They ratified his black belt on February 5, 1964, but he had to relinquish his Yoseikan grade}\\\cmidrule(lr){2-3}
                                      &     Q\&A          &       \makecell[l]{Question: What do organizations hold?\\
                                                                               Answer: Organizations hold competitions (tournaments) from local to international level.}\\\cmidrule(lr){2-3}
                                      &     Q\&A          &       \makecell[l]{Question: What is a useful tool to understand a kata?\\
                                                                               Answer: Bunkai}\\
                                                                    
    \bottomrule
    \end{tabular}
    % }
% \vspace{-1em}
  \caption{\textbf{Qualitative comparison under a realistic retrieval failure scenario.} We present a representative open-domain QA example where retrieved evidence is semantically related but weakly aligned with the true query intent, a common failure mode in deployed RAG systems. For clarity, we show the top-2 retrieved contexts for both standard RAG and \myalg, along with the feedback question-answer patches retrieved from~\autoref{eq:patchrag_sim}. Note that \hl{yellow highlight} indicates the answer for the given question, and it indicates that contexts or question-answer pairs are relevant to the user query $\mathrm{q}$.
  }
  \label{tab:bad_scenario}
  \vspace{-1em}
\end{table*}
\setlength{\tabcolsep}{6pt}

\subsection{Failure Scenario}
We analyze a challenging failure case that is representative of realistic deployed RAG systems, where user feedback is available but retrieval remains ambiguous due to semantically adjacent yet intent-misaligned evidence. In this scenario, even with the dual intent--context retrieval strategy described in~\autoref{eq:patchrag_sim}, some retrieved question-answer pairs are not directly aligned with the target query intent.

As illustrated in~\autoref{tab:bad_scenario}, this setting reflects a common practical condition in open-domain QA, where relevant information is scattered across related topics and retrieved contexts may partially overlap with, but not fully resolve, the user’s underlying intent. Despite this ambiguity, we observe that the feedback patches retrieved by \myalg consistently contain stronger intent-aligned signals than those surfaced by standard RAG, resulting in evidence that is more actionable for correcting the model’s behavior.
This qualitative analysis highlights that \myalg does not rely on idealized retrieval conditions. Instead, it remains effective under realistic retrieval noise, demonstrating its robustness in scenarios where feedback is imperfect and evidence relevance is only weakly aligned with the query.

\section{Acknowledgements}
We utilized Large Language Models (\eg ChatGPT) to assist in refining the textual content and to generate feedback on this work. These tools were used solely for writing support, and were not employed for code generation or implementation.

% \section{Example Appendix}
% \label{sec:appendix}

% This is an appendix.

\end{document}